\pdfoutput=1

\documentclass[11pt]{article}

\usepackage[preprint]{acl}

\usepackage{times}
\usepackage{latexsym}

\usepackage[T1]{fontenc}

\usepackage[utf8]{inputenc}

\usepackage{microtype}

\usepackage{inconsolata}

\usepackage{graphicx}

\usepackage{subcaption}
\usepackage{multirow}
\usepackage{bbding}
\usepackage{pifont}
\usepackage{extpfeil}
\usepackage{makecell}
\usepackage{booktabs}

\title{LLaVA-MLB: Mitigating and Leveraging Attention Bias\\for Training-Free Video LLMs}

\author{Leqi Shen$^{1,2}\footnotemark[4]\footnotemark[1]$
\quad
Tao He$^{4,5}\footnotemark[1]$
\quad
Guoqiang Gong$^{3}$\footnotemark[2]
\quad
Fan Yang$^{1,2}$
\quad
Yifeng Zhang$^{3}$
\\
\textbf{Pengzhang Liu$^{3}$}
\quad
\textbf{Sicheng Zhao$^{2}$}\footnotemark[3]
\quad
\textbf{Guiguang Ding$^{1,2}$}\footnotemark[3]\\
$^{1}$ School of Software, Tsinghua University
\quad
$^{2}$ BNRist, Tsinghua University \\
$^{3}$ JD.com
\quad
$^{4}$ GRG Banking Equipment Co., Ltd.
\quad
$^{5}$ South China University of Technology \\
}

\begin{document}
\maketitle

{
\renewcommand{\thefootnote}{\fnsymbol{footnote}}
\footnotetext[4]{Work done during an internship at JD.com.}
\footnotetext[1]{Equal contribution.}
\footnotetext[2]{Project leader.\quad $^{\ddagger}$ Corresponding authors.}
}

\begin{abstract}
Training-free video large language models (LLMs) leverage pretrained Image LLMs to process video content without the need for further training. A key challenge in such approaches is the difficulty of retaining essential visual and temporal information, constrained by the token limits in Image LLMs. To address this, we propose a two-stage method for selecting query-relevant tokens based on the LLM attention scores: compressing the video sequence and then expanding the sequence. However, during the compression stage, Image LLMs often exhibit a positional attention bias in video sequences, where attention is overly concentrated on later frames, causing early-frame information to be underutilized. To alleviate this attention bias during sequence compression, we propose Gridded Attention Pooling for preserving spatiotemporal structure. Additionally, we introduce Visual Summarization Tail to effectively utilize this bias, facilitating overall video understanding during sequence expansion. In this way, our method effectively \textbf{M}itigates and \textbf{L}everages attention \textbf{B}ias (\textbf{LLaVA-MLB}), enabling the frozen Image LLM for detailed video understanding. Experiments on several benchmarks demonstrate that our approach outperforms state-of-the-art methods, achieving superior performance in both efficiency and accuracy. Our code will be released.
\end{abstract}

\section{Introduction}
\label{sec:intro}

The Video LLMs leverage large language models~\cite{achiam2023gpt, chiang2023vicuna, taori2023stanford, touvron2023llama, touvron2023llama2} (LLMs) for video-language reasoning, which can interpret video content and natural language instructions to generate contextually appropriate responses.
Generally, Video LLMs can be categorized into two main types: training-based and training-free.
Training-based methods are fine-tuned on large-scale video datasets, allowing the models to adapt to the temporal nature of video content. 
However, these methods typically require the collection and annotation of extensive supervised fine-tuning datasets, along with substantial computational resources for training.
In contrast, training-free methods directly leverage pretrained Image LLMs to interpret video content, without additional training. These methods improve the generalization performance of Image LLMs when applied to video data.

\begin{figure}[t]
    \centering
    \scalebox{1.0}
    {
    \includegraphics[width=\linewidth]{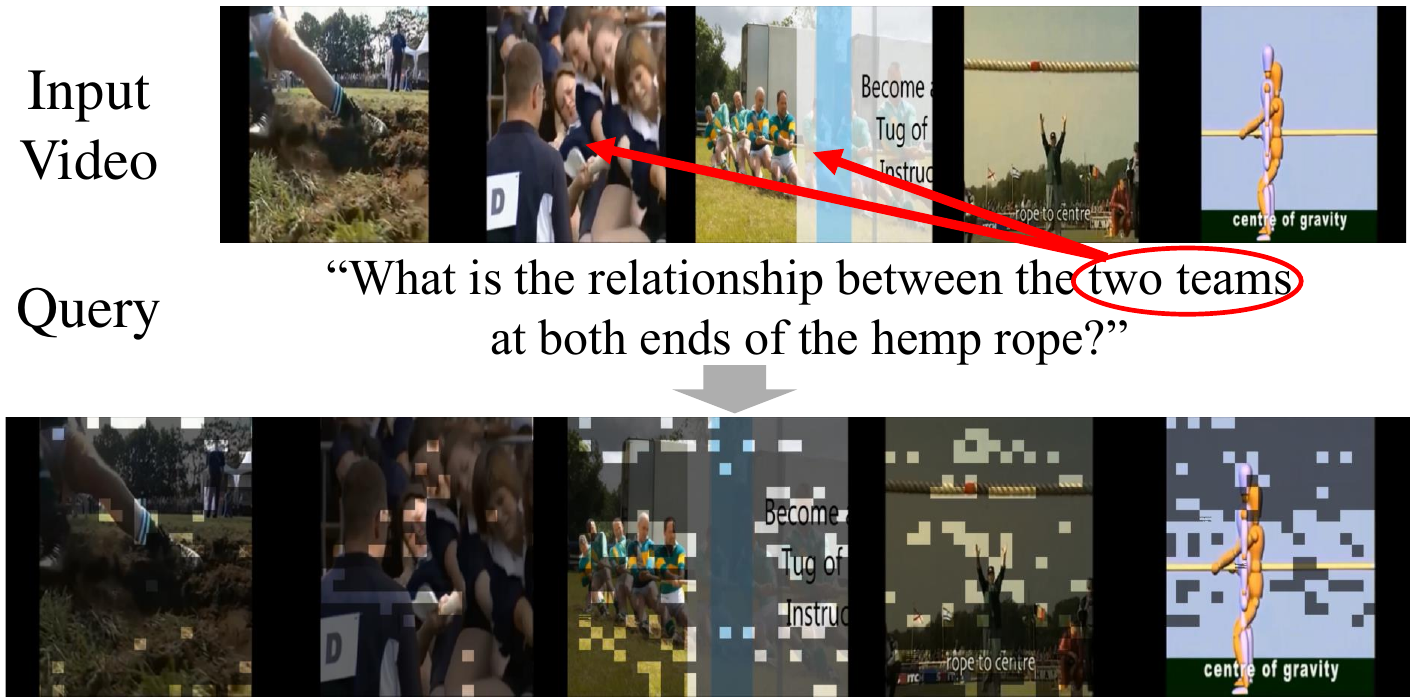}
    }
    \caption{An example from ANet-QA~\cite{yu2019activitynet} illustrates the positional attention bias. Out of the 2880 tokens from 5 sampled frames ($5\times 24 \times 24$), the 720 tokens with the highest attention scores are mostly concentrated in the last frame.}
    \label{fig:intro_vis_example}
\end{figure}

In these training-free methods, tokens from all video frames are fed together into the LLM.
Due to the LLM token limits, Image LLMs struggle to process a large number of tokens per frame and maintain a long temporal context.
To address this, we aim to input only the tokens that are relevant to the query.
Specifically, we propose a two-stage framework: sequence compression and sequence expansion. First, we use the attention scores from the LLM to select query-relevant tokens within a video segment. Second, to further enhance the temporal context, we merge multiple compressed segments in the temporal dimension. These compressed tokens are then fed into the LLM to generate the final output.
This would enable more fine-grained and accurate token compression, thereby effectively expanding the spatiotemporal dimensions. 

\begin{figure}[t]
    \centering
    \begin{subfigure}{1.0\linewidth} 
        \includegraphics[width=\linewidth]{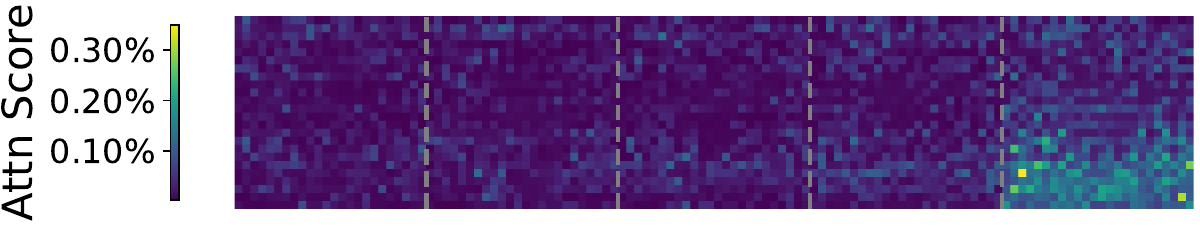} 
        \vspace{-17pt}
        \caption{
        5 sampled frames in sequential order.}
        \vspace{-2pt}
        \label{fig:intro_vis_hw}
    \end{subfigure}
    \begin{subfigure}{1.0\linewidth} 
        \includegraphics[width=\linewidth]{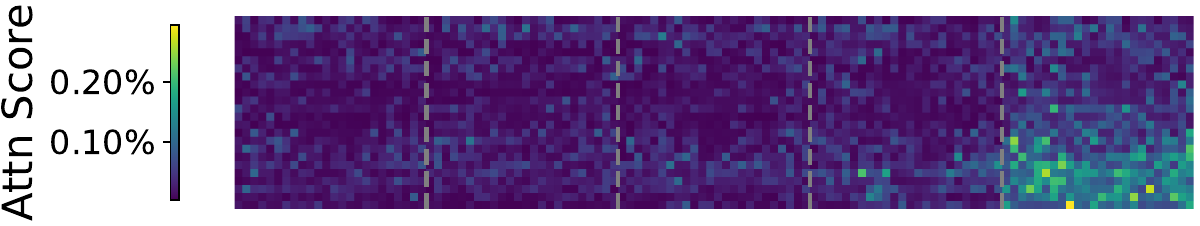} 
        \vspace{-17pt}
        \caption{
        5 sampled frames in reverse order.
        }
        \vspace{-2pt}
        \label{fig:intro_vis_hw_reverse}
    \end{subfigure}
    \begin{subfigure}{1.0\linewidth} 
        \includegraphics[width=\linewidth]{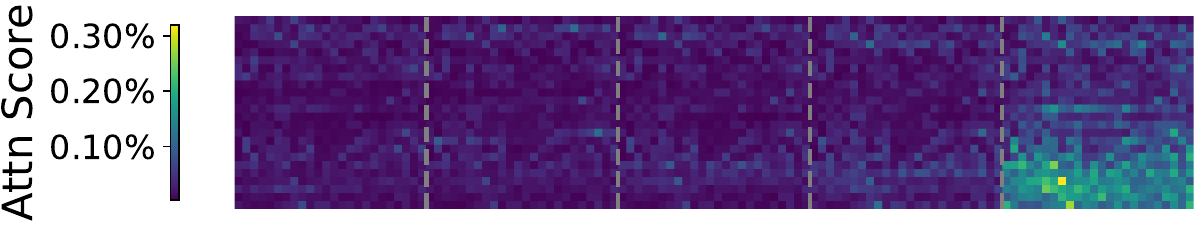} 
        \vspace{-17pt}
        \caption{
        A single frame repeated five times.
        }
        \vspace{-2pt}
        \label{fig:intro_vis_hw_single}
    \end{subfigure}
    \caption{
    The average attention scores on ANet-QA. For each video, tokens from 5 sampled frames ($5\times 24 \times 24$) are input into the LLM. The attention score is computed using the last query token at $3^{\rm th}$ layer. Brighter areas correspond to higher attention values. (a) shows five sequential frames, and (b) shows them in reverse. In (c), a single frame is sampled and repeated five times.
    }
    \label{fig:intro_vis}
\end{figure}

However, when compressing the sequence, we have found that a positional attention bias is evident at the tail of video sequences in Image LLMs.
As shown in Figure~\ref{fig:intro_vis_example}, the example video showcases a complex multi-scene tug-of-war event. Although the query relates to earlier ``two teams'' scenes, the Image LLM incorrectly focuses high attention on the last, irrelevant frame.
Additionally, Figure~\ref{fig:intro_vis_hw} quantifies the attention scores across the entire ANet-QA. 
We observe that most of the high attention tokens in the video are concentrated in the last frame, which is closer to the query.  
Out of the 2880 tokens from five sampled frames, we select the top 720 tokens. The average number of tokens retained in the last frame is 358.4, accounting for \textbf{49.8}\% of the total selected tokens. 
Furthermore, similar attention biases are observed in Figure~\ref{fig:intro_vis_hw_reverse} and \ref{fig:intro_vis_hw_single}, confirming that such a bias is consistent across different sequence compositions. 
This suggests that positional attention bias at the end of video sequences is inherent in Image LLMs, independent of the input’s semantic content.

Such attention bias negatively affects query-relevant token selection during sequence compression.
If we retain video tokens with high attention scores in this manner, 
it becomes difficult to preserve information that is farther from the query. 
Potentially useful information from earlier frames may be discarded and not utilized by the model, limiting the overall effectiveness.

To mitigate the attention bias inherent in video sequences for Image LLMs, 
we tackle the problem from two distinct angles, effectively \textbf{M}itigating and \textbf{L}everaging attention \textbf{B}ias (\textbf{LLaVA-MLB}).
First, we minimize the negative impact of attention bias during sequence compression. Second, we leverage the properties of attention bias for sequence expansion.

Specifically,
on the one hand, 
to mitigate the concentration of selected high attention tokens towards the tail, we introduce a \textit{Gridded Attention Pooling} module. This approach selects the most relevant visual tokens to the query within multiple local grids, ensuring that the selected visual tokens are evenly distributed across multiple frames in the spatiotemporal dimension. This method not only avoids the bias issue but also preserves the spatiotemporal structure of the visual tokens.  
On the other hand, the phenomenon of attention scores concentrating on the tail can be further leveraged for better global video understanding.
To this end, we propose a \textit{Visual Summarization Tail} module, which uses a large kernel pooling operation to generate a small token set that represents global summary information. These tokens are placed at the end of the visual sequence, allowing the query to focus more on the global context.

We evaluate our LLaVA-MLB across several tasks, including text generation~\cite{maaz2023video}, open-ended VideoQA~\cite{yu2019activitynet, li2016tgif}, and multiple-choice VideoQA~\cite{xiao2021next, mangalam2023egoschema}.
Our method outperforms existing approaches across all benchmarks. Compared to the previous SOTA model, SF-LLaVA~\cite{xu2024slowfast}, our LLaVA-MLB achieves \textbf{0.8}\% higher accuracy on ANet-QA and \textbf{4.2}\% higher accuracy on EgoSchema, while requiring only \textbf{57}\% of the pre-filling time.

Our contributions are summarized as follows: 

\noindent $\bullet$ We identify a positional attention bias phenomenon in Image LLM for video understanding, where the attention scores between the query and the video are more concentrated on the tail-end visual tokens.

\noindent $\bullet$ We propose the \textit{Gridded Attention Pooling} module to mitigate the bias phenomenon for enhanced token compression, and the \textit{Visual Summarization Tail} module to leverage this bias phenomenon for better capturing global video information.

\noindent $\bullet$ We validate the effectiveness of the proposed method on multiple video understanding benchmarks, demonstrating superior performance that exceeds current state-of-the-art methods, in terms of both efficiency and effectiveness.

\section{Related Work}
\label{sec:related_work}

\noindent \textbf{Video LLMs.}
Building on advancements in LLMs~\cite{achiam2023gpt, chiang2023vicuna, taori2023stanford, touvron2023llama, touvron2023llama2, ouyang2022training} and Image LLMs~\cite{team2023gemini, liu2024improved, liu2024llavanext, li2023blip, alayrac2022flamingo, li2024llava, liu2024visual, dai2023instructblipgeneralpurposevisionlanguagemodels}, researchers explore Video LLMs~\cite{zhang2023video, cheng2024videollama, maaz2023video, lin2023video, song2024moviechat, song2024moviechat+, li2023videochat, ma2023vista, xu2024pllava, zhang2024llavanextvideo} for diverse applications in video-language comprehension.  
Video-LLaMA~\cite{zhang2023video} enables video comprehension by combining frozen encoders and frozen LLMs with a Video Q-former.
Video-ChatGPT~\cite{maaz2023video} introduces a new video instruction dataset for adapting the image-based LLaVA model to video tasks.
MovieChat~\cite{song2024moviechat} implements an effective memory management mechanism for long videos.
Chat-UniVi~\cite{jin2024chat} applies multiscale dynamic visual tokens to represent images or video tokens. 
PLLaVA~\cite{xu2024pllava} introduces a post-training method, adapting pretrained Image LLMs for dense video understanding.
LLaVA-NeXT-Video~\cite{zhang2024llavanextvideo} further fine-tunes image-based LLaVA-Next on video data, with LLaVA-Next-Video-DPO~\cite{zhang2024llavanextvideo} adding direct preference optimization.
Nevertheless, most video LLMs rely on extensive fine-tuning with large-scale video datasets, which incurs significant computational costs.

\begin{figure*}[t!]
  \centering
  \scalebox{0.95}
  {
  \includegraphics[width=\linewidth]{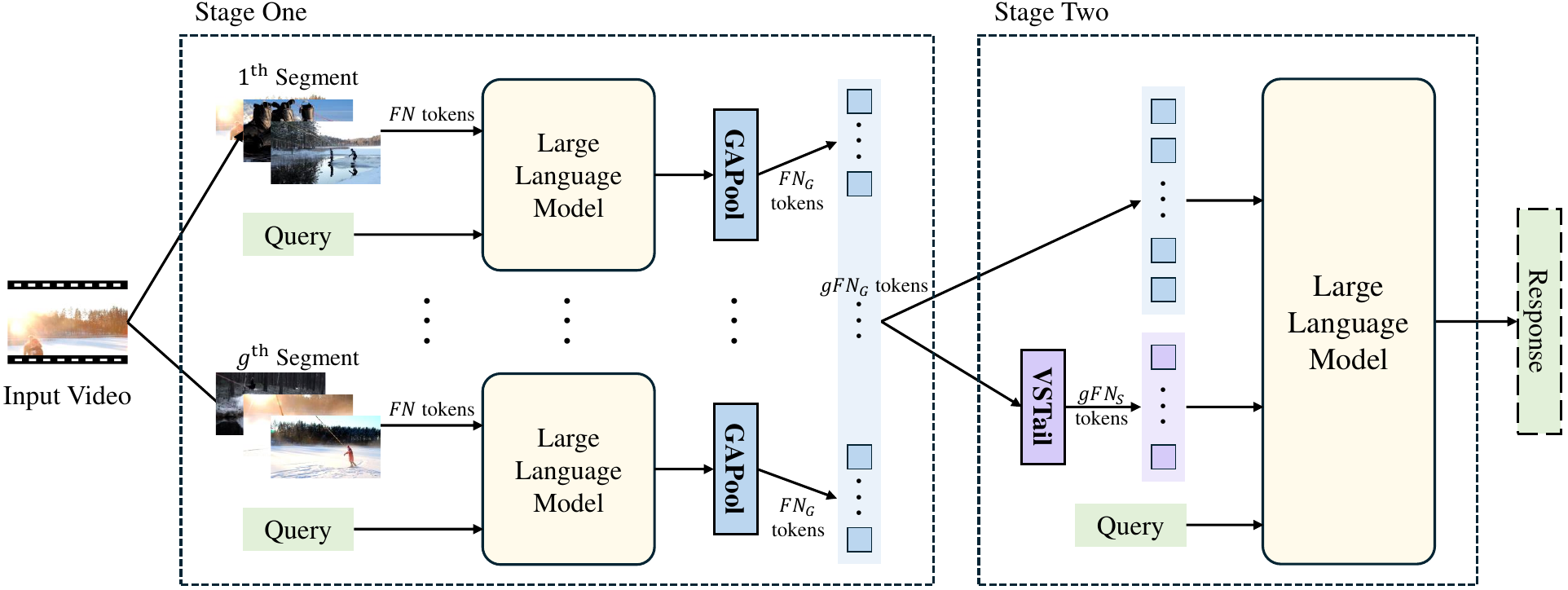}
  }
   \caption{Illustration of LLaVA-MLB. We employ the two-stage SegAttn framework to extend the visual sequence. GAPool is introduced to mitigate attention bias, improving token compression. VSTail leverages attention bias to enhance overall video understanding.}
   \label{fig:method_framework}
\end{figure*}

\noindent \textbf{Training-Free Video LLMs.}
By directly applying pretrained Image LLMs without additional training, training-free Video LLMs~\cite{kim2024image, wu2024freeva, han2024free, xu2024slowfast, wang2024videotree, zhang2023simple, wang2024videoagent} present a cost-efficient option for video understanding. These training-free methods can be categorized into two types: caption-based~\cite{wang2024videotree, zhang2023simple, wang2024videoagent} and vision-based~\cite{kim2024image, wu2024freeva, han2024free, xu2024slowfast}. Caption-based methods convert videos into a series of densely sampled frame captions, allowing LLMs to respond to text queries based on these captions. However, these methods are limited by the quality of the caption generation process. 
In this paper, we focus on vision-based methods, which rely on pretrained LLMs to process sampled frames directly for query response. 
FreeVA~\cite{wu2024freeva} explores different temporal pooling aggregation strategies on video features. 
SF-LLaVA~\cite{xu2024slowfast} employs a two-stream architecture to aggregate video features through pooling.
Although these query-agnostic pooling methods reduce the token count per frame, allowing for longer temporal context, they significantly reduce spatial detail.
To mitigate this, we aim to compress tokens by selecting those most relevant to the query.

Recent Free Video-LLM~\cite{han2024free} is also a query-conditioned method that uses query prompt information to guide the selection of regions of interest in the video. Our LLaVA-MLB differs significantly from Free Video-LLM. 
First, Free Video-LLM computes relevance scores using the global text features extracted from the CLIP text encoder, raising concerns about the granularity and accuracy of its computations.
In contrast, our LLaVA-MLB employs attention scores generated by advanced LLMs. 
Second, rather than focusing on query-relevant regions, LLaVA-MLB focuses on query-relevant token-level information, achieving a more granular focus on critical video details.

\noindent \textbf{Token Compression for Vision LLMs.}
Token compression has emerged as a crucial technique for improving the efficiency of vision LLMs. 
MobileVLM~\cite{chu2023mobilevlm} designs an efficient projector between visual and text features, reducing the inference budget.
LLaVA-Prumerge~\cite{shang2024llava} exploits the sparsity observed in the visual encoder to dynamically select the most crucial visual tokens.
FastV~\cite{chen2024image} proposes a versatile plug-and-play method for pruning visual tokens in LMM layers.
LOOK-M~\cite{wan2024look} focus utilizes a text-prior technique to compress the KV cache.

However, these token compression are generally restricted by the limited frame length of input video sequences. To address this limitation, we propose a two-stage solution, involving both sequence compression and expansion, that enables the effective processing of longer video sequences using frozen Image LLMs.

\section{Methodology}
\label{sec:methods}

\subsection{Preliminary}

The rapid advancements in Image LLMs stem from large-scale image-text datasets and image instruction data. However, extending this success to video requires extensive training on large video datasets, which is resource-intensive. Training-free Video LLMs provide an efficient alternative by leveraging pretrained Image LLMs without additional training.

Given a video, we first uniformly sample $T$ frames. Each frame is independently processed by an image encoder, such as CLIP-L~\cite{radford2021learning}. The generated tokens are subsequently mapped into the language space through a pretrained projection layer. This process yields $N=H \times W$ tokens per frame.
These tokens from all sampled frames are concatenated to form the complete video input for the LLM, with a resolution of $T \times H \times W$. 
Finally, both the visual input $\mathbf{V}$ and the textual query prompt $\mathbf{Q}$ are input to the LLM to generate a response.

However, high token resolutions $H \times W$ and large sampled frame length $T$ may exceed the LLM's maximum token limit. This challenge demands a careful trade-off between spatial detail and temporal span. Previous methods~\cite{wu2024freeva, xu2024slowfast, han2024free} attempt to reduce the token resolution using pooling, enabling longer frame sequences. 
However, the pooling operation significantly compromises spatial details. 
In contrast, we propose LLaVA-MLB, which employs query-relevant tokens to represent the input video, effectively preserving spatiotemporal detail with fewer tokens.

\subsection{Segmented Attention Framework}

In contrast to Free Video-LLM~\cite{han2024free}, which relies on the CLIP text encoder to obtain query-relevant regions, our LLaVA-MLB selects query-relevant tokens with high attention scores from powerful LLMs.
Traditional token pruning methods, however, only shorten the visual sequence, lacking the flexibility to expand it for temporal information. 
Therefore, we propose Segmented Attention Framework (SegAttn), a two-stage approach in Figure~\ref{fig:method_framework}: sequence compression in the first stage and sequence expansion in the second stage. To simplify the process, we employ a strategy that compresses by a certain factor $g$ and expands by the same amount.

Furthermore, as shown in Figure~\ref{fig:intro_vis}, a significant attention bias appears at the end of the video input, which distorts the selection of query-relevant tokens. To mitigate this bias, we propose Gridded Attention Pooling (GAPool) to preserve spatio-temporal structure. Alternatively, to leverage this bias, we propose Video Summarization Tail (VSTail) to enhance overall understanding.

Specifically, in stage one, $T$ sampled frames are divided into $g$ segments, with each segment containing $F$ frames. For each segment, the $F$ frames combined with the query prompt are input to the LLM to generate attention scores. Based on these scores, GAPool compresses $FN$ tokens into $FN_G$ tokens, where $N_G=\frac{N}{g}$. 

\begin{table*}[t]
\centering
\resizebox{\linewidth}{!}
{
\begin{tabular}{c|c|cc|ccc}
\toprule
\multirow{2}{*}{Method} & \multirow{2}{*}{\# Input Tokens} & \multicolumn{2}{c|}{Pre-filling Time} & VCGBench  & ANet-QA        & EgoSchema \\
                        &                                  & Parallel         & Sequential         & Avg Score & Accuracy/Score & Accuracy  \\ \hline
SF-LLaVA~\cite{xu2024slowfast}                & $3680$                             & \multicolumn{2}{c|}{0.9659 s}           & 3.04      & 55.5/\textbf{3.4}       & 47.2      \\ \hline
LLaVA-MLB$^{G}$             & $2880$                             & \textbf{0.5473} s           & \textbf{0.7352} s             & 3.06      & \textbf{56.3}/\textbf{3.4}       & \textbf{51.4}      \\
LLaVA-MLB$^{G+S}$               & $2880+240$                          & 0.5701 s           & 0.7579 s             & \textbf{3.11}      & 55.6/3.3       & \textbf{51.4}      \\ \bottomrule
\end{tabular}
}
\caption{Efficiency comparisons using 7B LLMs. 
\# Input Tokens denotes the number of tokens fed into the LLM for response generation.
The pre-filling time is the time the model takes to generate the first output token.
Parallel and Sequential refer to the compression of stage one conducted in parallel and sequentially, respectively.
LLaVA-MLB$^{G}$ denotes SegAttn with GAPool. LLaVA-MLB$^{G+S}$ denotes SegAttn with both GAPool and VSTail, where VSTail introduces 240 additional summary tokens.}
\label{tab:ablation_speed}
\end{table*}

In stage two, compressed tokens from all segments are aggregated, yielding a total of $FN=gFN_G$ tokens. Moreover, VSTail appends $gFN_S$ summary tokens at the end of the sequence, where $N_S \ll N$. The final response is generated from these tokens and the query prompt.

\subsection{Gridded Attention Pooling}

In stage one, we obtain attention scores from the pre-filling phase of the LLM. We leverage the attention scores from the shallow layers to identify tokens strongly associated with the query. Shallow layer scores are beneficial for two reasons:
(1) According to~\cite{chen2024image,wang2023label}, 
attention distribution is typically more balanced in shallow layers, where information is assembled to form semantic representations. Therefore, high attention scores in shallow layers help to capture query-relevant tokens.
(2) Shallow layers are computationally efficient, as they avoid the need to propagate through the entire LLM.
In particular, we utilize attention scores from the $k^{\rm th}$ layer’s last query token, where $k=3$. 

Due to attention bias, high attention scores often concentrate on tokens at the end of visual sequences. To mitigate this issue, we propose GAPool to select the most relevant visual tokens within multiple local grids, which maintains the sequence's spatiotemporal structure. Specifically, 
GAPool divides $FN$ tokens of each segment into grids:
\begin{align}
    F \times N & = F \times H \times W \nonumber \\
     &= F \times (H_{G}G_h) \times (W_{G}G_w) \nonumber \\
    &= (FH_{G}W_{G}) \times G_h \times G_w \nonumber \\
    &= (FN_G) \times G_h \times G_w ,
\end{align}
where $G_h$ and $G_w$ denote the grid height and weight.
$H_G$ and $W_G$ denote the number of grids along the height and width. $N_G$ is the total number of grids per frame. From each grid, GAPool selects the query-relevant token with the highest attention score:
\begin{gather}
    (p_*, q_*) = \underset{1 \leq p \leq G_h, \, 1 \leq q \leq G_w}{\operatorname{arg\,max}} \, \mathbf{A}_{p q}^{G_{fij}}
\end{gather}
where $\mathbf{A}^{G_{fij}}$ denotes the attention scores of the grid located at the $i^{\rm th}$ row and $j^{\rm th}$ column in the $f^{\rm th}$ frame. In $G_{fij}$, we output only the token at position $(p_*, q_*)$. 
Therefore, we effectively compress the segment sequence into $FN_G$ tokens, 
corresponding to a resolution of $F \times H_{G} \times W_{G}$,
achieving a compression rate of $g = G_{h} \times G_{w}$.

To simplify the sequence expansion in stage two, the sequence expansion factor is set to match the compression factor of each segment sequence.
We concatenate the compressed tokens from all segments, totaling $FN=gF\frac{N}{g}=gFN_G$ tokens.
Finally, these compressed tokens from $T=gF$ sampled frames are combined with a query prompt to produce the final response in the LLM.

\subsection{Video Summarization Tail}

In contrast to GAPool, which mitigates attention bias in the selection of query-relevant tokens, VSTail leverages the concentration of attention at the visual input’s tail. 
VSTail implements a summary branch that applies large-kernel pooling to the stage one output, creating a small token set as the new tail for the visual input. This tail corresponds to regions with high attention scores, enhancing the overall understanding of the video.

Specifically, on the $gF \times H_G \times W_G$ tokens from stage one, we apply average pooling with a large kernel size defined as $S_{h} \times S_w$. Each frame is downsampled from $H_{G} \times W_{G}$ to $H_{S} \times W_{S}$, where $H_{S} = \frac{H_{G}}{S_{h}}, W_{S} = \frac{W_{G}}{S_{w}}$. Consequently, the summary sequence has $gFN_S = gF \times H_S \times W_S$ tokens, where $N_S \ll N$. Each token represents a receptive field of size $P_h \times P_w = (G_{h}S_{h}) \times (G_{w}S_{w})$ in the original input. The appended sequence at the visual input's tail serves as a summary of the input video.

\begin{table*}[t!]
\centering
\resizebox{\linewidth}{!}
{
\begin{tabular}{cc|c|c|cccccc}
\toprule
\multicolumn{2}{c|}{\multirow{2}{*}{Method}} & \multirow{2}{*}{\makecell{LLM\\Size}} & \multirow{2}{*}{\makecell{Vision\\Encoder}} & \multicolumn{6}{c}{VCGBench (Score)}    \\
\multicolumn{2}{c|}{}                        &                           &                                 & CI   & DO   & CU   & TU   & CO   & Avg  \\ \hline
\multicolumn{1}{c|}{\multirow{8}{*}{\makecell{Training-based\\Video LLMs}}} & Video-LLaMA~\cite{zhang2023video}             & 7B                        & CLIP-G                          & 1.96 & 2.18 & 2.16 & 1.82 & 1.79 & 1.98 \\
\multicolumn{1}{c|}{} & Video-LLaMA2~\cite{cheng2024videollama}            & 7B                        & CLIP-L                          & 3.16 & 3.08 & 3.69 & 2.56 & 3.14 & 3.13 \\
\multicolumn{1}{c|}{} & Video-ChatGPT~\cite{maaz2023video}           & 7B                        & CLIP-L                          & 2.50 & 2.57 & 2.69 & 2.16 & 2.20 & 2.42 \\
\multicolumn{1}{c|}{} & MovieChat~\cite{song2024moviechat}               & 7B                        & CLIP-G                          & 2.76 & 2.93 & 3.01 & 2.24 & 2.42 & 2.67 \\
\multicolumn{1}{c|}{} & VideoChat~\cite{li2023videochat}               & 7B                        & CLIP-G                          & 2.23 & 2.50 & 2.53 & 1.94 & 2.24 & 2.29 \\
\multicolumn{1}{c|}{} & Vista-LLaMA~\cite{ma2023vista}              & 7B                        & CLIP-G                          & 2.44 & 2.64 & 3.18 & 2.26 & 2.31 & 2.57 \\
\multicolumn{1}{c|}{} & LLaVA-NeXT-Video~\cite{zhang2024llavanextvideo}        & 7B                        & CLIP-L                          & 3.39 & 3.29 & 3.92 & 2.60 & 3.12 & 3.26 \\
\multicolumn{1}{c|}{} & LLaVA-NeXT-Video-DPO~\cite{zhang2024llavanextvideo}    & 7B                        & CLIP-L                          & 3.64 & 3.45 & 4.17 & 2.95 & 4.08 & 3.66 \\ \hline \hline
\multicolumn{1}{c|}{\multirow{4}{*}{\makecell{Training-free\\Video LLMs}}} & IG-VLM~\cite{kim2024image}                  & 7B                        & CLIP-L                          & 3.11 & 2.78 & 3.51 & 2.44 & 3.29 & 3.03 \\
\multicolumn{1}{c|}{} & SF-LLaVA~\cite{xu2024slowfast}                & 7B                        & CLIP-L                          & 3.09 & 2.70  & 3.57 & \textbf{2.52} & 3.35 & 3.04 \\ 
\multicolumn{1}{c|}{} & \textbf{LLaVA-MLB$^{G}$}               & 7B                        & CLIP-L                          & 3.18 & \textbf{2.85} & 3.55 & 2.45  & 3.29  & 3.06  \\
\multicolumn{1}{c|}{} & \textbf{LLaVA-MLB$^{G+S}$}               & 7B                        & CLIP-L                          & \textbf{3.22} & 2.83 & \textbf{3.61} & \textbf{2.52}  & \textbf{3.38} & \textbf{3.11}    \\
\bottomrule
\end{tabular}
}
\caption{Results of text generation using 7B LLMs. 
The upper part shows methods trained on video data, while the lower part displays training-free Video LLMs.
For the training-free method, the best metric is highlighted in bold.}
\label{tab:sota_vcg}
\end{table*}

\begin{table*}[t!]
\centering
\resizebox{\linewidth}{!}
{
\begin{tabular}{cc|c|c|cc|cc}
\toprule
\multicolumn{2}{c|}{\multirow{3}{*}{Method}}  & \multirow{3}{*}{\makecell{LLM\\Size}} & \multirow{3}{*}{\makecell{Vision\\Encoder}} & \multicolumn{2}{c|}{Open-Ended VideoQA} & \multicolumn{2}{c}{Multiple Choice VideoQA} \\
\multicolumn{2}{c|}{}                         &                           &                                 & ANet-QA            & TGIF-QA     & NExTQA              & EgoSchema                \\ 
\multicolumn{2}{c|}{}                         &                           &                                 & \multicolumn{2}{c|}{Accuracy/Score}            & \multicolumn{2}{c}{Accuracy}             \\ \hline
\multicolumn{1}{c|}{\multirow{11}{*}{\makecell{Training-based\\Video LLMs}}} & Video-LLaMA~\cite{zhang2023video}             & 7B                        & CLIP-G                          & 12.4/1.1           & -                  & -                    & -                      \\
\multicolumn{1}{c|}{}     &  Video-LLaMA2~\cite{cheng2024videollama}            & 7B                        & CLIP-L                          & 50.2/3.3           & -                  & -                    & 51.7                  \\
\multicolumn{1}{c|}{}     & Video-ChatGPT~\cite{maaz2023video}           & 7B                        & CLIP-L                          & 35.2/2.7           & 51.4/3.0           & -                    & -                     \\
\multicolumn{1}{c|}{}     & Video-LLaVA~\cite{lin2023video}             & 7B                        & ViT-L                           & 45.3/3.3           & 70.0/4.0           & -                    & -                      \\
\multicolumn{1}{c|}{}     & MovieChat~\cite{song2024moviechat}              & 7B                        & CLIP-G                          & 45.7/3.4           & -                  & -                    & -                      \\
\multicolumn{1}{c|}{}     & MovieChat+~\cite{song2024moviechat}              & 7B                        & CLIP-G                          & 48.1/3.4           & -                  & 54.8                & -                      \\
\multicolumn{1}{c|}{}     & Video-Chat~\cite{li2023videochat}              & 7B                        & CLIP-G                          & 26.5/2.2           & 34.4/2.3           & -                    & -                      \\
\multicolumn{1}{c|}{}     & Vista-LLaMA~\cite{ma2023vista}             & 7B                        & CLIP-G                          & 48.3/3.3           & -                  & 60.7                & -                      \\
\multicolumn{1}{c|}{}     & PLLaVA~\cite{xu2024pllava}                  & 7B                        & CLIP-L                          & 56.3/3.5           & 77.5/4.1           & -                    & -                      \\
\multicolumn{1}{c|}{}     & LLaVA-NeXT-Video~\cite{zhang2024llavanextvideo}        & 7B                        & CLIP-L                          & 53.5/3.2           & -                  & -                    & -                      \\
\multicolumn{1}{c|}{}     & LLaVA-NeXT-Video-DPO~\cite{zhang2024llavanextvideo}    & 7B                        & CLIP-L                          & 60.2/3.5           & -                  & -                    & -                      \\ \hline \hline
\multicolumn{1}{c|}{\multirow{6}{*}{\makecell{Training-free\\Video LLMs}}} & FreeVA~\cite{wu2024freeva}                  & 7B                        & CLIP-L                          & 51.2/\textbf{3.5}           & -                  & -                   & -                     \\
\multicolumn{1}{c|}{}     &IG-VLM~\cite{kim2024image}                  & 7B                        & CLIP-L                          & 54.3/3.4           & 73.0/4.0           & 63.1                & 35.8                  \\
\multicolumn{1}{c|}{}     &Free Video-LLM~\cite{han2024free} & 7B                        & CLIP-L                          & 54.8/3.4           & 77.8/4.1           & -                   & -                     \\
\multicolumn{1}{c|}{}     &SF-LLaVA~\cite{xu2024slowfast}                & 7B                        & CLIP-L                          & 55.5/3.4           & 78.7/\textbf{4.2}           & 64.2                & 47.2                  \\ 
\multicolumn{1}{c|}{}     & \textbf{LLaVA-MLB$^{G}$}               & 7B                        & CLIP-L                          & \textbf{56.3}/3.4                           & 78.6/4.1                & 67.3                 & \textbf{51.4}                 \\ 
\multicolumn{1}{c|}{}     & \textbf{LLaVA-MLB$^{G+S}$}              & 7B                        & CLIP-L                          & 55.6/3.3                  & \textbf{78.8}/4.1                  & \textbf{67.5}                   & \textbf{51.4}                     \\
\bottomrule
\end{tabular}
}
\caption{Results of open-ended VideoQA and multiple choice VideoQA using 7B LLMs.}
\label{tab:sota_mutliopen}
\end{table*}

\subsection{Efficiency Analysis}
\label{sec:efficiency}

In our framework, these hyper-parameters directly influence the model's efficiency, $F$, $k$, $g=G_h \times G_w$, and $S_h \times S_w$.

$F$ denotes the number of frames per segment. 
In stage one, the attention score is computed from the $k^{\rm th}$ layer, enabling the LLM to stop forward propagation at this early layer, thereby saving significant computational time.
In stage two, the sequence expansion factor is set equal to the compression factor $g$ from stage one, where $g$ is determined by the GAPool's grid size $G_h \times G_w$.
In VSTail, the pooling kernel size $S_h \times S_w$ determines the number of generated summary tokens.

Furthermore, in stage one, we are able to apply parallel compression to multiple segments, effectively reducing inference time. When sufficient hardware resources are available, a parallel strategy can be adopted. However, when only a single GPU is available, the sequential strategy remains efficient, as stage one only requires a shallow LLM, which does not result in significant time overhead.

In Table~\ref{tab:ablation_speed}, we conduct an efficiency comparison between our method and the state-of-the-art SF-LLaVA~\cite{xu2024slowfast}. 
Both SF-LLaVA and our method are built upon LLaVA-NeXT~\cite{liu2024llavanext}.
With just \textbf{57}\% pre-filling time, LLaVA-MLB$^{G}$ surpasses SF-LLaVA by \textbf{0.8}\% accuracy on ANet-QA and by \textbf{4.2}\% on EgoSchema.
Although implemented as a two-stage framework, our proposed LLaVA-MLB surpasses previous approaches in both performance and efficiency. 

\section{Experiments}
\label{sec:experiments}

We conduct evaluations across three tasks: text generation, open-ended VideoQA, and multiple choice VideoQA. 
We introduce benchmarks and metrics for each task individually and implementation details in Appendix~\ref{sec:appendix_data}.

\begin{table}[t]
\centering
\resizebox{\linewidth}{!}{
\begin{tabular}{c|c|cc|c}
\toprule
\multirow{2}{*}{\makecell{Segment\\Length $F$}} & \multirow{2}{*}{\makecell{\# Tokens\\$gFN_G$}}  & \multicolumn{2}{c|}{Pre-filling Time} & ANet-QA       \\
                                  &                                   & Parallel & Sequential & Accuracy/Score  \\ \hline
3                                 & 1728                                  & 0.3514 s  & 0.4788 s & 55.0/3.3    \\
4                                 & 2304                                 & 0.4519 s & 0.6071 s & 55.2/3.3      \\
~5$^{*}$                                 & 2880                                   & 0.5473 s & 0.7352 s &  56.3/\textbf{3.4}       \\
6                                 & 3456                                 & 0.6686 s & 0.9018 s  & \textbf{56.8}/\textbf{3.4}    \\ 
\bottomrule
\end{tabular}
}
\caption{Ablation study on segment length $F$ in LLaVA-MLB$^{G}$. \# tokens denotes the final visual input length, calculated as the sum of the compression tokens across all segments.}
\label{tab:ablation_segattn}
\end{table}

\begin{table}[t]
\centering
\resizebox{\linewidth}{!}
{
\begin{tabular}{c|cc|c}
\toprule
The $k^{\rm th}$ Layer  & \multicolumn{2}{c|}{Pre-filling Time}  & ANet-QA        \\
Attention Score    & Parallel & Sequential  & Accuracy/Score \\ \hline
 2       & 0.5459 s & 0.7105 s  & 55.3/3.3           \\
~3$^{*}$       & 0.5473 s & 0.7352 s    & \textbf{56.3}/\textbf{3.4}           \\
 5       & 0.5694 s & 0.8040 s   & 55.8/\textbf{3.4}           \\
 8      & 0.5986 s & 0.9064 s      & 56.0/\textbf{3.4}        \\
 10      & 0.6188 s & 0.9735 s    & 55.9/\textbf{3.4}           \\
\bottomrule
\end{tabular}
}
\caption{Ablation study on the $k^{\rm th}$ layer attention score in LLaVA-MLB$^{G}$. We use attention score from the $k^{\rm th}$ layer for GAPool.}
\label{tab:ablation_prunelayer}
\end{table}

\subsection{Comparisons with State-of-the-Arts}

In Table~\ref{tab:sota_vcg}, we report the performance of the text generation task evaluated on VCGBench using 7B LLMs. 
LLaVA-MLB$^{G+S}$ shows significant improvement over LLaVA-MLB$^{G}$ in CU, TU, and CO, confirming the effectiveness of VSTail, which appends a summary sequence to the visual input tail to enhance overall video understanding.
Compared to prior state-of-the-art methods, our model achieves comparable performance in CU, TU, and CO. 
Notably, our model achieves significant improvements in both CI and DO. 
Compared to IG-VLM, LLaVA-MLB$^{G+S}$ improves the CI score by \textbf{0.11}\%, and LLaVA-MLB$^{G}$ enhances the DO score by \textbf{0.07}\%, highlighting our model's superior ability to capture fine-grained details in video content.

Table~\ref{tab:sota_mutliopen} presents a comparison between our LLaVA-MLB and previous training-free methods on both open-ended VideoQA and multiple choice VideoQA tasks, utilizing 7B LLMs.
For open-ended VideoQA, LLaVA-MLB$^{G}$ achieves comparable performance on TGIF-QA and increases accuracy by \textbf{0.8}\% on ANet-QA.
For multiple-choice VideoQA, LLaVA-MLB$^{G+S}$ demonstrates marked improvements, surpassing SF-LLaVA by \textbf{3.3}\% accuracy on NExTQA and \textbf{4.2}\% accuracy on EgoSchema.

In Appendix~\ref{sec:appendix_34B}, we show the results with 34B LLMs, which confirm our model's robustness and generalizability.

\begin{figure}[t]
    \centering
    \scalebox{0.97}
    {
    \includegraphics[width=0.85\linewidth]{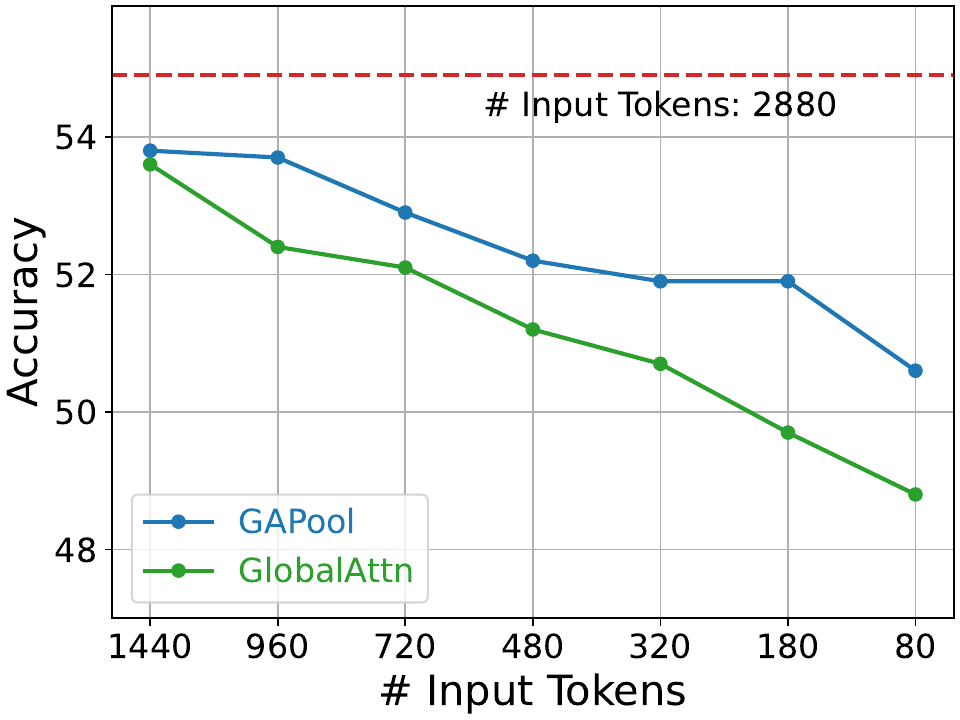} 
    }
    \caption{Ablation study on GAPool in LLaVA-MLB$^{G}$ using a 5-frame segment on ANet-QA.
    The red dashed line indicates the accuracy without compression.
    The number of the tokens fed into stage two is denoted by \# input tokens.
    In GAPool, 1440, 960, 720, 480, 320, 180, and 80 indicate grid sizes of 1$\times$2, 1$\times$3, 2$\times$2, 2$\times$3, 3$\times$3, 4$\times$4, and 6$\times$6. }
    \label{fig:ablation_gapool_anet}
\end{figure}

\begin{table}[t]
\centering
\resizebox{\linewidth}{!}{
\begin{tabular}{c|c|cc|c}
\toprule
Grid Size & \# Frames & \multicolumn{2}{c|}{Pre-filling Time} & ANet-QA     \\
$G_h \times G_w$  & $g \times F$   & Parallel & Sequential      & Accuracy/Score  \\ \hline
$1 \times 2$                     & 10       & 0.4419 s  & 0.5044 s & 55.8/3.3                     \\
~$2 \times 2^{*}$                     & 20       & 0.5473 S & 0.7352 S & \textbf{56.3}/\textbf{3.4}                      \\
$2 \times 3$                     & 30    & 0.6940 S & 1.0072 S   & 56.1/\textbf{3.4}                     \\
$2 \times 4$                     & 40     & 0.7800 S  & 1.2175 S  & 54.6/3.3                     \\ \bottomrule
\end{tabular}
}
\caption{Ablation study on GAPool in LLaVA-MLB$^{G}$. \# Frames denotes the total sampled frames $g \times F = (G_{h}G_{w}) \times F$.}
\label{tab:ablation_GAPoolinSegAttn}
\end{table}

\begin{table*}[t]
\centering
\resizebox{\linewidth}{!}
{
\begin{tabular}{c|cc|c|ccc}
\toprule
\multirow{2}{*}{\makecell{Pooling Size\\$S_h \times S_w$}} & \multirow{2}{*}{\makecell{Receptive Field\\$P_h \times P_w$}} & \multirow{2}{*}{\makecell{Frame Size\\$H_S \times W_S$}} & \multirow{2}{*}{\makecell{\# Input Tokens\\$gFN_{G} + gFH_{S}W_{S}$}} & VCGBench  & ANet-QA        & EgoSchema \\
                             &                             &                             &                                  & Avg Score & Accuracy/Score & Accuracy  \\ \hline
- & - & -                                                                & $2880+0$                           & 3.06     & \textbf{56.3}/\textbf{3.4}                & 51.4       \\  \hline
$12 \times 12$ & $24 \times 24$ & $1 \times 1$                                                                & $2880+20$                           & \textbf{3.12}     & 54.6/3.3      & 51.8       \\
$6 \times 6$ & $12 \times 12$ & $2 \times 2$                                                                & $2880+80$                           & \textbf{3.12}     & 55.1/3.3      & 51.4       \\
$4 \times 4$ & $8 \times 8$ & $3 \times 3$                                                                & $2880+180$                           & 3.10     & 55.2/3.3      & \textbf{52.0}       \\
~$3 \times 4^{*}$ & $6 \times 8$ & $4 \times 3$                                                                & $2880+240$                           & 3.11     & 55.6/3.3      & 51.4       \\
$3 \times 3$ & $6 \times 6$ & $4 \times 4$                                                                & $2880+320$                           & 3.09     & 55.0/3.3      & 50.8       \\
$2 \times 3$ & $4 \times 6$ & $6 \times 4$                                                                & $2880+480$                           & 3.08     & 55.1/3.3      & 51.4      \\
$2 \times 2$ & $4 \times 4$ & $6 \times 6$                                                                & $2880+720$                           & 3.06     & 55.8/\textbf{3.4}      & 51.0       \\
\bottomrule
\end{tabular}
}
\caption{Ablation study on VSTail's pooling size in LLaVA-MLB$^{G+S}$. Receptive field is determined by the GAPool grid size and the VSTail pooling size, $P_h \times P_w = (G_{h}S_{h}) \times (G_{w}S_{w})$. Frame size denotes the token size per frame in the VSTail summary sequence, $H_S \times W_S = \frac{H}{P_h} \times \frac{W}{P_w}$. \# input tokens consists of $gFN_{G}=gF\frac{N}{g}$ compressed tokens from stage one and $gFH_{S}W_{S}$ tokens from VSTail.}
\label{tab:ablation_vstail}
\end{table*}

\subsection{Ablation Study on SegAttn Framework}

\noindent \textbf{Effect of Segment Length $F$.}
In Table~\ref{tab:ablation_segattn}, we evaluate our LLaVA-MLB$^{G}$ with various segment lengths. 
The frame number per segment $F$ determines the total sampled frames $gF$ from the input video. 
Increasing the sequence length provides more temporal information from the videos, leading to improved performance on ANet-QA as the frame length increases. 
Using $gF = (G_{h}G_{w})F= 4\times 6 = 24$ sampled frames, we achieve a notable accuracy of \textbf{56.8}\% on ANet-QA. 
However, a greater frame number implies more tokens for the LLM to process, which increases inference time. 
To balance efficiency and accuracy, we set $F$ to 5.

\noindent \textbf{Effect of the $k^{\rm th}$ Layer Attention Score.}
Table~\ref{tab:ablation_prunelayer} shows the effect of attention score from different layers in LLaVA-MLB$^{G}$. As $k$ grows, additional layers are required for the LLM inference in stage one, resulting in longer inference times. The third shallow layer obtains the best performance (\textbf{56.3}\% Accuracy) with high efficiency, leading us to select $k=3$.

\subsection{Ablation Study on GAPool}

\noindent \textbf{Effect of using a single segment.}
In Figure~\ref{fig:ablation_gapool_anet}, we conduct experiments in LLaVA-MLB$^{G}$ using a single 5-frame segment on ANet-QA.
In the case of a single segment, we extract the compression tokens in stage one and input them directly into the LLM in stage two, without merging tokens from additional segments.
In GlobalAttn, the tokens with the top-$\frac{2880}{G_{h}G_{w}}$ attention scores are selected as the final input into the LLM.
Across various compression rates, GAPool consistently outperforms GlobalAttn by better selecting query-relevant tokens. 
Our grid-based token selection strategy preserves the spatiotemporal structure of video tokens, effectively mitigating attention bias during sequence compression.

\noindent \textbf{Effect of Grid Size $G_{h} \times G_{w}$.}
Table~\ref{tab:ablation_GAPoolinSegAttn} presents the impact of GAPool’s grid size $G_{h} \times G_{w}$ in LLaVA-MLB$^{G}$. In the two-stage SegAttn, the temporal dimension is expanded in alignment with the compression ratio $g = G_{h}G_{w}$. We find that when the sampled frame number surpasses 20, accuracy begins to degrade with increasing frames. This is likely due to the information loss incurred by higher compression ratios. A lower compression rate preserves detail, while a higher frame number introduces more temporal information, both of which need careful balancing. The optimal \textbf{56.3}\% accuracy is achieved with a grid size of $G_{h} \times G_{w} = 2 \times 2$.

\subsection{Ablation Study on VSTail}

\noindent \textbf{Effect of Pooling Size $S_{h} \times S_{w}$.}
Table~\ref{tab:ablation_vstail} presents an ablation study on the pooling size $S_{h} \times S_{w}$ in VSTail for LLaVA-MLB$^{G+S}$. A larger pooling size enlarges the receptive field of the summary token, leading to a smaller frame size in the summary sequence. As a result, each token in the sequence conveys a more comprehensive summary. 
Although VSTail shows limitations on ANet-QA, using one summary token per frame yields the highest average score of \textbf{3.12}\% on VCGBench. 
As pooling size decreases, VCGBench performance also declines, potentially due to an attention bias toward the visual input's tail. As depicted in Figure~\ref{fig:intro_vis}, the final $\sim$300 tokens exhibit a significantly higher attention score, suggesting that the optimal length of the summary sequence should ideally not exceed 300 tokens. Given VSTail’s performance on various datasets, we set $S_{h} \times S_{w}= 3 \times 4$, resulting in 240 summary tokens.

In Appendix~\ref{sec:appendix_Spos}, we further analyze the impact of VSTail’s summary sequence position.

\section{Conclusion}

In this paper, we present a novel approach LLaVA-MLB for training-free video understanding in large language models, particularly focusing on mitigating attention bias in video sequences for Image LLMs. 
To mitigate the concentration of selected high attention tokens towards the tail, we propose Gridded Attention Pooling to preserve the spatiotemporal structure of video sequences.
Alternatively, we propose Visual Summarization Tail to leverage this bias for better global video understanding.
In this way, our LLaVA-MLB effectively selects query-relevant tokens, enabling detailed video comprehension in frozen Image LLMs.
Extensive experiments validate the effectiveness of LLaVA-MLB on various video understanding tasks.
Our model surpasses current training-free approaches in both efficiency and accuracy.

\section*{Limitations}

\noindent \textbf{Query-relevant Token Selection.}
Although GAPool effectively selects the most relevant token within each non-overlapping grid, it fails to capture all tokens associated to the query-relevant object. This limitation arises from the inherent attention bias in image LLMs, which causes the model to prioritize tokens at the end of the video sequence, hindering the selection of query-relevant object tokens. To mitigate this issue, our GAPool ensures that each grid contains a token with a high attention score, thereby preserving the spatiotemporal structure.

In addition, Free Video-LLM~\cite{han2024free} selects query-relevant regions by incorporating an additional text encoder, which introduces extra parameters. However, due to discrepancies between the text encoder and the image LLM, its performance is inferior (see Table~\ref{tab:sota_mutliopen}).

\noindent \textbf{Lightweight Fine-tuning vs. Training-free Methods.}
While lightweight fine-tuning from image LLMs to video can improve performance with minimal resources, it requires maintaining multiple specialized model checkpoints. In contrast, our LLaVA-MLB leverages a single pre-trained image LLM for both image and video understanding without modification. This makes our method particularly attractive for edge devices and real-time applications where efficiency is critical.

\noindent \textbf{Attention Bias.}
As highlighted in Section~\ref{sec:intro}, attention bias is an inherent challenge in image LLMs for video understanding. Since training-free video LLMs preserve the original image LLM weights, this bias remains unaltered in the final processing stage. Our LLaVA-MLB approach does not eliminate the bias but instead mitigates and leverages it to construct video input sequences with rich spatiotemporal information. To directly address this issue, fine-tuning could be a valuable direction for future research.

\bibliography{custom}

\clearpage

\appendix

\newcommand{\tabfootnotesize}{\fontsize{8}{9}\selectfont}

\makeatletter
  \newcommand\figcaption{\def\@captype{figure}\caption}
  \newcommand\tabcaption{\def\@captype{table}\caption}
\makeatother

\begin{minipage}[b]{\textwidth}
\centering
\resizebox{0.95\linewidth}{!}{
\begin{tabular}{c|c|c|cc|cc|c}
\toprule
\multirow{3}{*}{Method} & \multirow{3}{*}{\makecell{LLM\\Size}} & \multirow{3}{*}{\makecell{Vision\\Encoder}} & \multicolumn{2}{c|}{Open-Ended VideoQA} & \multicolumn{2}{c|}{Multiple Choice VideoQA} & \multirow{2}{*}{VCGBench} \\
                        &                           &                                 & ANet-QA            & TGIF-QA            & NExTQA              & EgoSchema              &                           \\
                        &                           &                                 & \multicolumn{2}{c|}{Accuracy/Score}     & \multicolumn{2}{c|}{Accuracy}                & Avg Score                 \\ \hline
Video-LLaMA2~\cite{cheng2024videollama}            & 46.7B                     & CLIP-L                          & 50.3/3.4           & -                  & -                   & 53.3                   & 3.15                      \\
PLLaVA~\cite{xu2024pllava}                  & 34B                       & CLIP-L                          & 50.9/3.7           & 80.6/4.3           & -                   & -                      & -                         \\
LLaVA-NeXT-Video~\cite{zhang2024llavanextvideo}        & 34B                       & CLIP-L                          & 58.8/3.4           & -                  & -                   & -                      & 3.34                      \\
LLaVA-NeXT-Video-DPO~\cite{zhang2024llavanextvideo}    & 34B                       & CLIP-L                          & 64.4/3.6           & -                  & -                   & -                      & 3.77                      \\ \hline \hline
IG-VLM~\cite{kim2024image}                  & 34B                       & CLIP-L                          & 58.4/\textbf{3.5}           & 79.1/4.2           & 70.9                & 53.6                   & 3.09                      \\
SF-LLaVA~\cite{xu2024slowfast}                & 34B                       & CLIP-L                          & 59.2/\textbf{3.5}           & 80.6/\textbf{4.3}           & 72.0                  & 55.8                   & 3.32                      \\ 
\textbf{LLaVA-MLB$^{G}$}              & 34B                       & CLIP-L                          & \textbf{59.4}/\textbf{3.5}                & \textbf{81.9}/4.2                 & 73.5                   & \textbf{57.4}                     & 3.34                       \\
\textbf{LLaVA-MLB$^{G+S}$}               & 34B                       & CLIP-L                          & 58.9/\textbf{3.5}                 & 81.3/4.2               & \textbf{73.6}                   & 56.8                      & \textbf{3.35}                         \\
\bottomrule
\end{tabular}
}
\tabcaption{Results of open-ended VideoQA, multiple choice VideoQA, and text generation using 34B LLMs.}
\label{tab:sota_34b}
\end{minipage}

\begin{minipage}[b]{0.45\textwidth}
\centering
\resizebox{1.0\linewidth}{!}{
\begin{tabular}{c|cccccc}
\toprule
\multirow{2}{*}{\makecell{VSTail\\Position}} & \multicolumn{6}{c}{VCGBench (Score)}    \\
                                 & CI   & DO   & CU   & TU   & CO   & Avg  \\ \hline
None                             & 3.18 & \textbf{2.85} & 3.55 & 2.45  & 3.29  & 3.06 \\
Head                             & \textbf{3.22}  & 2.83  & 3.55  & 2.41   & 3.32 & 3.07     \\
~Tail$^{*}$                             & \textbf{3.22} & 2.83 & \textbf{3.61} & \textbf{2.52}  & \textbf{3.38} & \textbf{3.11} \\
\bottomrule
\end{tabular}
}
\tabcaption{Ablation study on the VSTail's summary sequence position in LLaVA-MLB$^{G+S}$. None denotes no VSTail inclusion. 
Head and Tail refer to placing the summary sequence at the start and end of the visual input, respectively.}
\label{tab:ablation_vstailpos}
\end{minipage}

\section*{Appendix}

\section{Experimental Settings}
\label{sec:appendix_data}

\noindent \textbf{Text Generation.}
We evaluate video-to-text generation performance on VCGBench~\cite{maaz2023video}, a benchmark comprising 500 videos sampled from ActivityNet-1.3~\cite{yu2019activitynet}. Consistent with previous training-free methods~\cite{kim2024image, xu2024slowfast}, we use the GPT-3.5-Turbo-0125 model to assign relative scores to model outputs across five dimensions: \textbf{C}orrectness of \textbf{I}nformation (CI), \textbf{D}etail \textbf{O}rientation (DO), \textbf{C}ontextual \textbf{U}nderstanding (CU), \textbf{T}emporal \textbf{U}nderstanding (TU), and \textbf{CO}nsistency (CO). The average score across these metrics (Avg Score) provides an overall performance assessment.

\noindent \textbf{Open-ended VideoQA.}
The open-ended VideoQA task includes ActivityNet-QA (ANet-QA)~\cite{yu2019activitynet} and TGIF-QA~\cite{li2016tgif}, where Video LLMs generate open-ended answers based on both the video content and the query. These questions typically have single-word ground-truth answers. Following previous methods~\cite{kim2024image, wu2024freeva, han2024free, xu2024slowfast}, we employ the GPT-3.5-Turbo-0125 model to evaluate the responses in terms of accuracy (true or false) and quality (scored from 0 to 5).

\noindent \textbf{Multiple Choice VideoQA.}
Video LLMs are evaluated on their ability to select the correct answer 

~\\
~\\
~\\
~\\
~\\
~\\
~\\
~\\
~\\
~\\

\noindent from a set of predefined options. We evaluate performance using the NExTQA~\cite{xiao2021next} and EgoSchema~\cite{mangalam2023egoschema} benchmarks.

\noindent \textbf{Implementation Details}
Following recent training-free methods~\cite{kim2024image, han2024free, xu2024slowfast}, we employ the Image LLM, LLaVA-NeXT~\cite{liu2024llavanext}, for video understanding without additional training. Each sampled frame from the input video is resized to $336 \times 336$ and processed by the CLIP-L vision encoder~\cite{radford2021learning}, producing $H \times W = 24\times 24$ tokens. 
For a video, we sample $(G_{h}G_{w})\times F=4\times 5$ frames. In GAPool, attention scores are derived from the ${(k=3)}^{\rm th}$ layer and the grid size is $G_{h} \times G_{w} = 2\times 2$. The VSTail's pooling size is $S_{h} \times S_{w} = 3\times 4$. For further explanation, please see Section~\ref{sec:efficiency}.  
The two-stage SegAttn with GAPool is labeled LLaVA-MLB$^{G}$, while LLaVA-MLB$^{G+S}$ includes both GAPool and VSTail. When evaluating pre-filling times, Parallel denotes stage one executed in parallel across 4 A100 GPUs, while Sequential denotes it executed sequentially on a single A100 GPU.

\section{More Results}
\label{sec:appendix_results}

\subsection{Results with 34B LLMs}
\label{sec:appendix_34B}

Table~\ref{tab:sota_34b} shows the results with 34B LLMs, where our model outperforms prior approaches on all benchmarks. Specifically, our model improves accuracy by \textbf{1.3}\% on TGIF-QA, \textbf{1.6}\% on NExTQA, and \textbf{1.6}\% on EgoSchema. 
These consistent improvements across 7B and 34B LLMs confirm our model's robustness and generalizability.

\subsection{Effect of Summary Sequence Position.}
\label{sec:appendix_Spos}
Further validation of the VSTail's summary sequence position in LLaVA-MLB$^{G+S}$ is provided in Table~\ref{tab:ablation_vstailpos}.
Head and None perform similarly across all metrics, suggesting that summary tokens at the beginning do not improve video comprehension in Image LLMs. Conversely, when the summary sequence is placed at the end, Tail demonstrates marked improvements in CU, TU, and CO.  While CI and DO (Correctness of Information and Detail Orientation) focus on evaluating detailed video comprehension, CU, TU, and CO (Contextual Understanding, Temporal Understanding, and Consistency) assess overall understanding. Compared to None, Tail (VSTail) enhances global knowledge of video content while retaining fine-grained comprehension.

\end{document}